\documentclass[journal]{IEEEtran}
\usepackage[utf8]{inputenc}

\usepackage{algorithm}
\usepackage{algorithmic}
\usepackage{amsmath}
\usepackage{amssymb}
\usepackage{colortbl}
\usepackage{array}
\usepackage{bm}
\usepackage{booktabs}
\usepackage{cases}
\usepackage[pdftex]{graphicx}
\usepackage{indentfirst}
\usepackage{mathrsfs}
\usepackage{multirow}
\usepackage{url}
\usepackage{xcolor}
\usepackage{makecell}
\usepackage{pifont}
\usepackage{color}  
\usepackage{subfigure}
\usepackage{underscore}
\usepackage{soul}
\usepackage{float}
\usepackage{arydshln}
\usepackage[colorlinks,
            linkcolor=blue,
            anchorcolor=black,
            citecolor=black]{hyperref}

\newcommand{\ie}{\textit{i.e.},~}

\begin{document}

\title{Generative-based Fusion Mechanism for Multi-Modal Tracking}
\author{Zhangyong~Tang,
        Tianyang~Xu,
        Xuefeng~Zhu,
        Xiao-Jun~Wu$^*$,
        and Josef Kittler 
\thanks{Z. Tang, T. Xu, X.-J. Wu (Corresponding Author) and X. Zhu are with the School of Artificial Intelligence and Computer Science, Jiangnan University, Wuxi, P.R. China. (e-mail: \{zhangyong\_tang\_jnu; tianyang\_xu; xuefeng\_zhu95; xiaojun\_wu\_jnu\}@163.com)}

\thanks{J. Kittler is with the Centre for Vision, Speech and Signal Processing, University of Surrey, Guildford, GU2 7XH, UK. (e-mail: j.kittler@surrey.ac.uk)}
}
%
%
%
%
%
\maketitle
\sloppy

\begin{abstract}
Generative models (GMs) have received increasing research interest for their remarkable capacity to achieve comprehensive understanding. 
However, their potential application in the domain of multi-modal tracking has remained relatively unexplored. 
In this context, we seek to uncover the potential of harnessing generative techniques to address the critical challenge, information fusion, in multi-modal tracking.
In this paper, we delve into two prominent GM techniques, namely, Conditional Generative Adversarial Networks (CGANs) and Diffusion Models (DMs). 
Different from the standard fusion process where the features from each modality are directly fed into the fusion block, we condition these multi-modal features with random noise in the GM framework, effectively transforming the original training samples into harder instances.
This design excels at extracting discriminative clues from the features, enhancing the ultimate tracking performance.
To quantitatively gauge the effectiveness of our approach, we conduct extensive experiments across two multi-modal tracking tasks, three baseline methods, and three challenging benchmarks. 
The experimental results demonstrate that the proposed generative-based fusion mechanism achieves state-of-the-art performance, setting new records on GTOT, LasHeR and RGBD1K.
Code will be available at \textcolor{blue}{\textit{https://github.com/Zhangyong-Tang/GMMT}}.
\end{abstract}


\section{Introduction}\label{sec:introduction}
Due to the strict demand for the robustness of tracking systems in real-world applications, such as surveillance \cite{surveillance} and unmanned driving \cite{unmanneddriving}, visual object tracking with auxiliary modality, named as multi-modal tracking, draws growing attention recently.
For instance, the thermal infrared (TIR) modality provides more stable scene perception in the nighttime \cite{dfat}, and the depth (D) modality provides 3-D perception against occlusions \cite{rgbd1k}. 
In other words, the use of auxiliary modality can complement the visible image in challenging scenarios.

Regarding this, a series of fusion strategies have been explored to aggregate the multi-modal information.
These strategies fall into two main categories based on the output of their fusion block.
The first category involves adaptive weighting strategies \cite{cbpnet, JMMAC,siamcda}, where the fusion block produces weights (scalars, vectors, or tensors) multiplied to features from each modality
In contrast, the second category focuses on embedded fusion methods \cite{mfDimp, adrnet, tfnet}, which generate fused features using dedicated modules.
While these methods differ in the way they produce fused results, their training processes are quite similar. 
They are trained offline using multi-modal datasets like RGBT234 \cite{rgbt234} and LasHeR \cite{lasher}, and, from a discriminative perspective, their tracking performance consequently depends on how well they match the training data.
Additionally, it's important to note that these fusion modules remain fixed during inference. 
They apply a consistent projection from input images to fused features across all scenarios during testing, without understanding the specific content of the current input images.

\begin{figure}[t]
\centering
\includegraphics[width=0.49\textwidth]{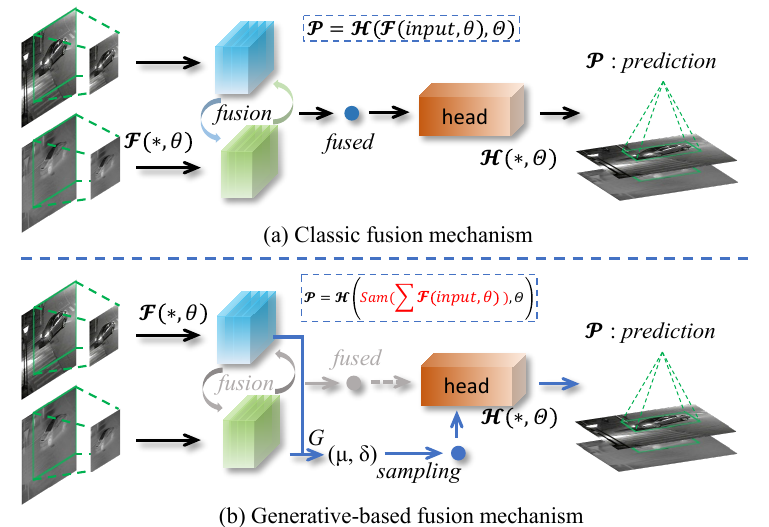}
\caption{Comparison between the original fusion mechanism and our generative-based fusion mechanism. GM is denoted as $G$, and we use $\mathcal{F}$ and $\mathcal{H}$ to refer to the feature extractor and task head, respectively.
The parameters associated with these two components are symbolized by $\theta$ and $\Theta$.}
\label{fig:paradigm}
\end{figure}

On the contrary, GMs have achieved great success due to their superiority in learning visual comprehension.
Accordingly, lots of downstream tasks have achieved promising performance, like image to image translation \cite{pix2pix} and multi-modal image fusion \cite{tgfuse}.
For example, in \cite{tgfuse}, a generator is employed to adaptively extract the salient clues among multiple inputs, and the discriminator further constrains the output to be high quality by reinforcing the textures globally.
However, extending its success on other multi-modal tasks to multi-modal tracking has not been sufficiently discussed yet.

Motivated by the aforementioned observations, the potential of applying the GMs to address the multi-modal information fusion is discussed in this paper, with a novel generative-based fusion mechanism being proposed for multi-modal tracking (GMMT), shown in Fig. \ref{fig:paradigm}.
In order to learn the external projection between the input and output, as well as the internal data distribution, the GMs require a longer training time and a bigger size of training data.
However, despite the emergence of several multi-modal datasets in recent years \cite{rgbd1k, lasher, hmft}, containing around 1000 videos captured across less than 1000 scenarios, there remains a significant diversity gap compared to widely-used datasets for training GMs, such as CelebA\cite{celebA}.
Therefore, the multi-modal feature pair grouped with a random factor is formulated as the input of the GMMT to avoid over-fitting.
Besides, to facilitate adaptive fusion of the certain image pair in the testing time, the original information from both modalities are retained as conditions.
Based on the above considerations, CGAN \cite{cgan} and DM \cite{ldm} are implemented in this paper. 
The generative-based fusion mechanism endows the fusion model with a better awareness of the noise, and thus the fused features are directional with less noise, as shown in Fig. \ref{fig:feature-difference}(c) and (d), which boosts the tracker to be a more accurate one. 
To validate the effectiveness of the proposed fusion mechanism, it is implemented on several RGB-T baseline trackers.
Consistent improvements can be obtained on all the evaluation metrics. 
Furthermore, extended experiments on the largest RGB-D benchmark \cite{rgbd1k} are also conducted to demonstrate the generalisation of the proposed fusion strategy.

In summary, our contributions can be summarised as follows:
\begin{itemize}
	\item We explore the potential of addressing the information fusion part of multi-modal tracking in a generative approach. To achieve this, a novel generative-based fusion mechanism is proposed, which boosts the fused features to be more discriminative. 
        \item A general fusion mechanism is proposed, with its generalisation demonstrated on multiple baseline methods, multiple benchmarks, and two multi-modal tracking tasks.
	\item Extensive experimental results demonstrate the proposed method as a state-of-the-art one in the RGB-T and RGB-D tracking fields.
\end{itemize}

\section{Related Work}\label{related work}
\subsection{RGB-T Trackers}
Before the access of extensive RGB-T datasets, including GTOT \cite{gtot}, RGBT210 \cite{RGBT210}, and RGBT234 \cite{rgbt234}, traditional RGB-T methods primarily relies on the sparse representation \cite{gtot} or handcrafted weighting strategies \cite{pixel-fusion2007} to tackle the information fusion task.
But these non-deep approaches suffer significant performance degradation in challenging scenarios.
As a result, recent research has been dominated by deep neural network-based fusion methods.
In recent studies, ranging from the simplest operation, concatenation \cite{mfDimp}, to the more complicated transformer architecture \cite{tbsi, vipt}, the researchers have tried various fusion strategies with multiple intentions, including learning modality importance \cite{JMMAC, taat}, reducing the multi-modal redundancy \cite{fusionnet, DAPNet}, propagating the multi-modal patterns\cite{cmpp}, learning the multi-modal prompts from the auxiliary modality \cite{vipt}, to name a few. 
With the increase in network complexity and the availability of larger training datasets, tracking results have gradually improved. 
This improvement has been particularly noteworthy since the release of the LasHeR \cite{lasher}, which promotes the development in a steep way.

\subsection{Generative Models}
While GMs have been one of the classical learning paradigms \cite{autoregressive}, they initially receive less attention during the early years of deep learning compared to the discriminative models. 
However, their significance is solidified after the introduction of GAN \cite{gan}.
GAN first showcase its prowess in image synthesis and subsequently is found to be a success in a range of tasks, including multi-modal image fusion \cite{tgfuse}, text-to-video generation \cite{videogeneration}, and text-to-audio generation \cite{audiogeneration}. 
After that, although more variations of GMs, such as variational auto-encoder \cite{vae}, and flow-based model \cite{flowbased}, also draw increasing attention, the downstream applications are still mainly based on GAN.

Until the proposal of the denoising diffusion model \cite{ddpm}, the interest in GAN falls gradually, as the diffusion model shows superior performance across multiple domains, notably excelling in visual-language generation \cite{textgeneration}.
Compared to GAN, the DM exhibits a more stable training procedure, and it can generate items in a more refined way.
However, it comes at the cost of significantly longer computation time.
Consequently, various research endeavors have been undertaken to address this computational challenge \cite{waveletdiffusion}.

\subsection{Generative Models Meet Tracking}
Several trackers have already explored the combination of GMs and tracking methods.
In the RGB tracking field, GMs are mainly introduced with two motivations, \ie generating more samples to improve the diversity \cite{sint++, ganpr2020, gantase2020},  and maintaining the most robust and long-lasting patterns \cite{vital, gantcsvt2019, ganpr2020}.
The generator is used for the first purpose while the second kind employs the discriminator to discard the less distinguishing patterns.

However, the application of generative models in the field of multi-modal tracking has received limited attention
As of now, only $\rm{BD^2}$Track \cite{BD2Track} explores the usage of recent generation models.
However, there are two main issues in this method.
Firstly, $\rm{BD^2}$Track employs a diffusion model to acquire fused classification features while retaining regression features learned discriminatively.
This configuration raises questions, as both classification and regression features within each modality encounter challenges and can potentially benefit from modality complementarity \cite{apfnet}.
Secondly, the effectiveness of the diffusion model in $\rm{BD^2}$Track lacks in-depth analysis, leaving space for a comprehensive understanding of its contribution.

In our design, we address the first issue by generating fused features before they are fed into the task heads, including both the classification and regression heads. 
This approach eliminates the need to handle classification and regression features separately within each modality, thus leveraging the inherent complementarity of modalities.
Furthermore, to tackle the second issue, we provide an intuitive explanation that verifies the superior performance of GMs. 
This explanation sheds light on why GMs are advantageous for multi-modal information fusion.
Notably, we extend our approach by implementing more than one type of GM. 
This inclusion enriches our discussion of applying GMs in multi-modal tracking, providing a more comprehensive exploration of this approach.
Additionally, we thoroughly evaluate the effectiveness of our proposed generative-based fusion mechanism across multiple baseline methods, benchmarks, and tracking tasks.

\section{Methodology}\label{method}
\subsection{Multi-modal Tracking}
Multi-modal tracking aims to obtain the prediction with the collaboration among multiple modalities, requiring the model to fuse relevant clues from the multi-modal input \{$\mathcal{X}^1$,...,$\mathcal{X}^m$\}. 
After pre-processing, the images are sent into the feature extractor and the fusion block.
However, these two blocks are sometimes entangled \cite{tbsi}, and therefore termed as $\mathcal{F}$ in combination.
The fused features $fused$ are then forwarded to the task head $\mathcal{H}$ to extract task-specific information.
Later, the final prediction $\mathcal{P}$ can be maintained after post-processing.
Since the pre- and post-processing procedures are not in the scope of this paper, they are skipped for brevity.
The mathematical description of multi-modal tracking is presented as follows:

\begin{equation}
    \centering \label{formulation:basic}
    \mathcal{P} = \mathcal{H}(\mathcal{F}(input, \theta), \Theta),
\end{equation}
where $\theta$ and $\Theta$ denote the learnable parameters of $\mathcal{F}$ and $\mathcal{H}$, respectively.
$input$ is the multi-modal image pair after pre-processing.

\subsection{Generative-based Fusion Mechanism}

\begin{figure}[t]
\centering
	\setcounter {subfigure} {0} a){
		\includegraphics[scale=0.5]{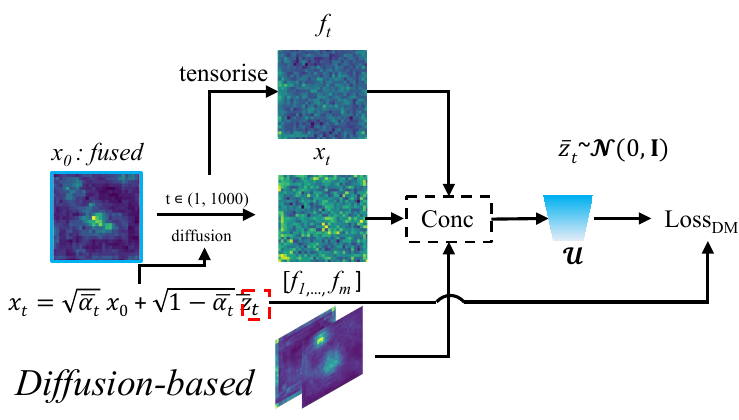}}
  \\
	\setcounter {subfigure} {0} b){
		\includegraphics[scale=0.45]{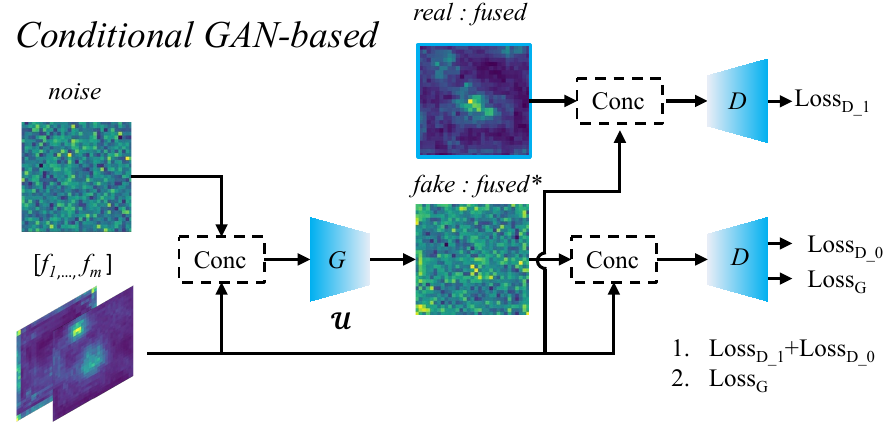}}
\caption{Illustration of the proposed GMMT.}
\label{fig:dmandcgan}
\end{figure}

Many multi-modal tasks have benefited from GMs, and this observation motivated us to explore the application of GMs in the field of multi-modal tracking.
To fulfill this objective, we introduce a novel fusion mechanism, termed GMMT, in this section.
Given that the fusion process is typically applied at the feature level, our GMMT is also thoughtfully designed and discussed within the embedding feature space.
In other words, the original fused features $fused$, the input of our GMMT, should be obtained beforehand, which aligns with our multi-stage training scheme.
Other than $fused$, the features from each modality ($f_{1},...,f_{m}$) should also be retained, which provides strong conditions to guide the fusion for the specific frame pair.
These analysis constrain the input of the GMs, but attach no limitation to the type of GMs.
Therefore, two popular GMs, \ie DM and CGAN, are involved in our method.

The DM-based GMMT is depicted in Fig. \ref{fig:dmandcgan}(a).
Following the DDIM \cite{ddim}, in the training stage, the original fused feature $fused$ serves as the $x_0$.
In the forward diffusion process, $x_0$ undergoes diffusion through the random Gaussian noise $\overline{z}_t$, as defined by the following formulation:
\begin{equation}
    \centering \label{formulation:dm}
    x_t = \sqrt{\overline{\alpha}_t}x_0 + \sqrt{1- \overline{\alpha}_t}\overline{z}_t,
\end{equation}
where the subscript $t$ is a random factor chosen from the interval [1, T], which defines how many steps $x_0$ are performed.
$\overline{\alpha}_t$ is a factorial of $\alpha_{1,..,t}$, which are the remainder of $\beta_{1,...,t}$.
Here $\beta_{t}$ is the predefined diffusion rate and determines how far the $t^{th}$ forward step goes.
Once the noisy representation $x_t$ is computed, the reverse diffusion process begins, aiming to recover a clean $x_{t-1}$.
It takes $x_t$ as input, along with ($f_{1},...,f_{m}$) as conditions, and the tensorised embedding of $t$, $f_t$, as a flag.
These elements are concatenated and fed into the U-shaped network $\mathcal{U}$.
$\mathcal{U}$ is then optimised by minimizing the $\mathcal{L}$2 loss between the output and noise $\overline{z}_t$, based on which the mean $\mu_{t-1}$ and variance $\sigma_{t-1}$ for the distribution of $x_{t-1}$ can be derived according to Eq. \ref{formulation:musigma}.
\begin{equation}
    \centering \label{formulation:musigma}
    \begin{aligned}
    \sigma_{t-1} & = \frac{1 - \overline{\alpha}_{t-1}}{1 - \overline{\alpha}_{t}} \beta_t ,\\
    \mu_{t-1} & = \frac{\sqrt{\alpha_t}(1 - \overline{\alpha}_{t-1})}{1 - \overline{\alpha}_{t}} x_t + \\
     & \frac{\sqrt{\overline{\alpha}_{t-1}}\beta_t}{1-\overline{\alpha}_t} \frac{1}{\sqrt{\overline{\alpha}_t}}(x_t - \sqrt{1 - \overline{\alpha}_t} \mathcal{U}(x_t, f_{1},...,f_{m}, f_t)).\\
    \end{aligned}
\end{equation}
Therefore, in the testing phase, the reverse diffusion process is executed iteratively, and in the end, the result can be sampled from the learned distribution of $x_{0}$.
But at the beginning, $x_t$ is replaced by random noise, and then the time flag $t$ is reversely traversed from T to 1.

In general, with the DM-based GMMT, the typical tracking process described in Eq. \ref{formulation:basic} develops to Eq.\ref{formulation:gmmt}:
\begin{equation}
    \centering \label{formulation:gmmt}
    \mathcal{P} = \mathcal{H}(Sam(\sum(\mathcal{F}(input, \theta))), \Theta),
\end{equation}
where $Sam$ is the abbreviation of the sampler, which means sampling data from the generated distribution.

The CGAN-based GMMT is displayed in Fig. \ref{fig:dmandcgan}(b).
Following the widely-used CGAN \cite{cgan}, the discriminator $D$ and generator $G$ are trained iteratively.
To train the $D$, the synthesised $fused^*$ and the original $fused$ are one-hot labelled, assigning 1 to $fused$ and 0 to $fused^*$. 
After that, separate losses are computed for $fused^*$ and $fused$, denoted as $Loss_{D_0}$ and $Loss_{D_1}$, respectively.
Aiming at distinguishing the real and fake data, $D$ is optimised by minimising $Loss_{D} = Loss_{D\_0} + Loss_{D\_1}$.
After training $D$, its parameters are frozen, and the learning process of $G$ commences.
$fused^*$ is sent into $D$, and the corresponding loss $Loss_{G}$ is obtained and minimised.
Since $G$ is designed to deceive and mislead $D$, $Loss_{G}$ is equivalent to $Loss_{D\_1}$.
Notably, the loss in this part is calculated by mean square error.
To ensure a fair comparison, the architecture of $G$ mirrors that of $\mathcal{U}$ employed in DM-based GMMT.
Besides, since only $G$ is employed during inference, the introduction of $D$ is remained in the supplementary material.

As observed, the output of the CGAN-based GMMT consists of fake features, signifying that the distribution is not explicitly learned.
Consequently, the overall tracking process remains similar to Eq. \ref{formulation:basic}.

\subsection{Multi-modal Trackers}\label{sec:trainingdetails}
The proposed GMMT is implemented on three RGB-T trackers \ie a self-designed Siamese tracker, the ViPT \cite{vipt}, and TBSI \cite{tbsi}, which implies that $m=2$ during application.
During the discussion of GMMT, the fused features $fused$ are assumed pre-defined, indicating that the baseline trackers should be pre-trained beforehand.
This necessitates two training stages: one to train the baseline method and another to train the proposed GMMT.

For the three selected baseline trackers, the first training stage consists of two primary steps: training the feature extractor and the fusion block.
As to the Siamese tracker, SiamBAN \cite{siamban} with a single region proposal network is trained for each modality.
A straightforward convolution-based fusion block is constructed and trained for multi-modal fusion.
In this fusion block, the multi-modal features are initially concatenated and then pass through a convolutional block.
Regarding ViPT \cite{vipt}, the feature extractor is pre-trained, but its fusion block is re-trained for multi-modal tracking.
As to TBSI \cite{tbsi}, we use both the publicly available feature extractor and fusion block.
A comprehensive description of the implementation details for these baseline trackers is provided in the supplementary material.

Our GMMT is trained during the second stage of our approach.
To provide a stable input to GMMT, the feature extractor and the original fusion block are frozen while training the GMMT.
Besides, to harmonise the fusion approach with the tracking task, a learnable tracking head is appended, which means the loss in this stage combines the generative loss and the tracking loss $Loss_{track}$ inherited from the baseline method:

\begin{equation}
    \centering \label{formulation:loss}
    Loss = Loss_{track} + \lambda * Loss_{gen},
\end{equation}
where $\lambda$ is a hyper-parameter used to balance the contribution of generative loss.

During the testing phase, the original fusion block is discarded, and the fused feature generated by GMMT serves as the input to the subsequent task head $\mathcal{H}$.
Further details are provided in the supplementary material.

\section{Experiments}\label{experiments}
\subsection{Implementation Details}
Our experiments are conducted on an NVIDIA RTX3090Ti GPU card.
Our GMMT is trained on the training split of LasHeR with the parameters optimised by the SGD optimiser.
The learning rate is warmed up from 0.001 to 0.005 in the first 20 epochs and subsequently reduces to 0.00005 for the remaining 80 epochs.
We set the value of T to 1000.
 
\subsection{Benchmarks and Metrics}
The effectiveness of GMMT is verified on GTOT \cite{gtot}, LasHeR \cite{lasher}, and RGBD1K \cite{rgbd1k} benchmarks.
In these benchmarks, precision rate (PR), success rate (SR), normalised precision rate (NPR), recall (RE), and F-score are employed for evaluation, whose detail introductions can be found in the supplementary material.



\subsection{Ablation Study}\label{sec:ablationstudy}
\begin{table}[t]
    \centering
    \caption{Ablation study on GTOT and LasHeR. RAW means the generative loss is replaced with a $\mathcal{L}2$ loss.}
    \setlength{\tabcolsep}{1.0mm}
{
    \begin{tabular}{ccccccccccc}
    \toprule
       Framework & Dataset & Method & PR$\uparrow$ & NPR$\uparrow$ & SR$\uparrow$  \\
    \midrule
        Siamese & GTOT & Base &84.0&-&67.0\\
        Siamese & GTOT & +RAW&81.9&-&67.4\\
        Siamese & GTOT & +GMMT(CGAN)&81.4&-&67.5\\
        Siamese & GTOT & +GMMT(DM)&85.7&-&69.3\\
    \midrule
        Siamese & LasHeR & Base &50.9&47.4&39.8\\
        Siamese & LasHeR & +RAW&51.8&50.7&42.5\\
        Siamese & LasHeR & +GMMT(CGAN)&54.1&49.3&41.9\\
        Siamese & LasHeR & +GMMT(DM)&57.1&53.0&44.9\\
    \bottomrule
    \end{tabular}
    \label{tab:results-abalation-study}
    }
\end{table}

In this section, we present the ablation study of our GMMT.
We denote GMMT(DM) when the embedded generative model is a diffusion model and GMMT(CGAN) when a conditional GAN is employed within GMMT.
Additionally, to demonstrate that the improvement is indeed attributed to the generative-based fusion mechanism rather than the larger fusion block embedded in GMMT, we conduct an experiment where the generative loss is removed. 
In this scenario, the network is trained using an $\mathcal{L}2$ loss between the $fused$ and the network output. 
This variant is denoted with a subscript $RAW$ in Table. \ref{tab:results-ablation-study}.
For fairness, the network architecture remains consistent for all the competitors.
Thus, the primary distinction among these competitors lies in the loss function, which is mathematically defined as follows:

\begin{align} \label{formulation:lossgen}
    Loss_{gen}=\left\{
    \begin{aligned}
    & Loss_{RAW} = \mathcal{L}2(fused, output); \\
    & Loss_{CGAN} = E_{x\sim p_g(x)}[logD(x, f_{rgb}, f_{tir})] \\
    & \quad + E_{z\sim p_n(z)}[log(1-D(G(z, f_{rgb}, f_{tir})))]; \\
    & Loss_{DM} = \mathcal{L}2(noise, output); \\
\end{aligned}
\right.
\end{align}
$Loss_{gen}$ denotes the loss function for the network $\mathcal{U}$, and it can be switched to multiple choices.
When trained with the direct $\mathcal{L}2$ loss between the network input and output, $Loss_{gen} = Loss_{RAW}$.
When trained following the GAN framework, $Loss_{gen} = Loss_{CGAN}$.
$Loss_{CGAN}$ is a widely-used loss in the research of GAN, and its detail can be found in \cite{cgan}.
$Loss_{DM}$ is activated when the diffusion model is employed.
In this configuration, the network predicts the noise, and thus, $Loss_{DM}$ calculates the $\mathcal{L}2$ loss between the output $output$ and the noise $noise$ used in the forward diffusion process.

\begin{figure}[t]
\centering
\includegraphics[width=1.0\linewidth]{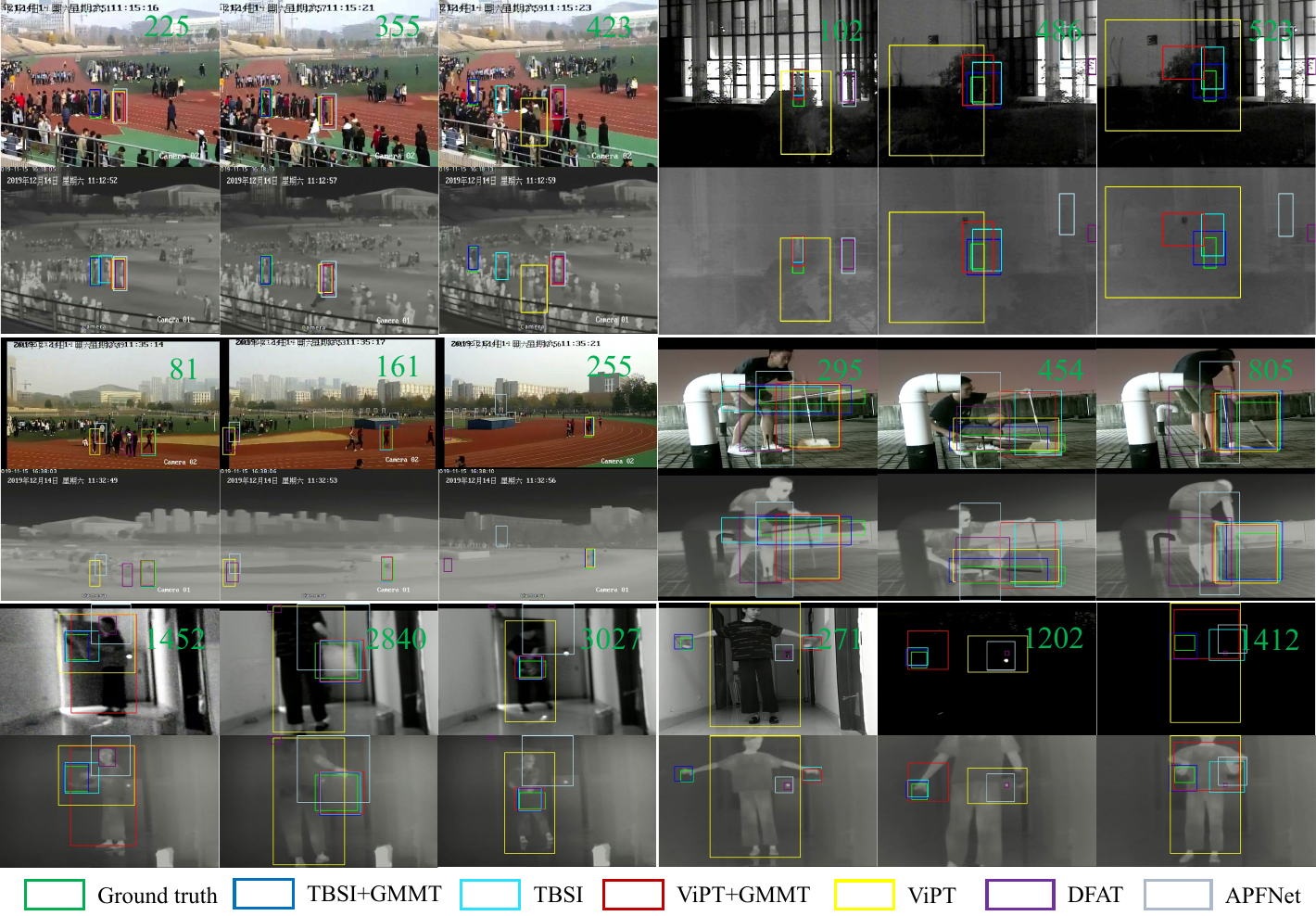}
\caption{Qualitative visualisation of several advanced RGB-T trackers. The exhibited image pairs are sampled from video $7rightorangegirl$, $10runone$, $ab\_bolstershaking$, $boyinplatform$, $besom3$, $ab\_rightlowerredcup\_quezhen$, which are introduced in a top-down and left-right way.}
\label{fig:visall}
\end{figure}

Table. \ref{tab:results-abalation-study} gives the results of the ablation study on GTOT and LasHeR benchmarks, using the Siamese-based tracker as the baseline method.
On LasHeR, the performance of the Siamese baseline is 39.8 on SR.
Replacing the fusion block with a larger network $\mathcal{U}$ results in an improvement to 42.5.
Later, when using the GMMT(CGAN), although a slight degradation of 0.6 appears on SR, a bigger enhancement of 2.3\% was maintained on PR, raised from 51.8 to 54.1.
Transitioning to the stronger generative model \cite{beatgan}, the DM, consistent improvements are obtained across all the metrics, reaching 57.1, 53.0, and 44.9.
Compared to the baseline tracker, significant gains of 6.2\%, 5.6\%, and 5.1\% are observed on PR, NPR, and SR, respectively.
It indicates that both the network with deeper architecture and our GMMT contribute to the promising performance. 
Similar conclusions are drawn from experiments on GTOT.
Compared to the baseline method, improvements of 1.7\% and 2.3\% are displayed in the results of GMMT(DM) on PR and SR.
Additionally, since GMMT(DM) performs better than GMMT(CGAN), the rest experiments are conducted based on DM.

\subsection{Compared with state-of-the-art Trackers}
We provide a quantitative comparison on LasHeR \cite{lasher}, RGBT234 \cite{rgbt234} and GTOT \cite{gtot}.

On LasHeR and RGBT234 benchmarks, several advanced RGB-T trackers are involved, including APFNet \cite{apfnet}, DMCNet \cite{DMCNet}, HMFT* \cite{hmft}, ProTrack \cite{protrack}, DFAT \cite{dfat}, HFRD \cite{hrfd}, ViPT* \cite{vipt}, TBSI \cite{tbsi}, and the ViPT* and TBSI modified by GMMT, termed as ViPT*+GMMT and TBSI+GMMT.
Here the superscript $*$ represents the results are reproduced by us.
As shown in Table. \ref{tab:results-lasher}, on LasHeR, the best results are obtained by TBSI+GMMT, reaching 70.7, 67.0 and 56.6 on PR, NPR, and SR, respectively. 
Compared to the original TBSI, GMMT improves the PR, NPR, and SR by 1.5\%, 1.3\%, and 1.0\%.
Combining GMMT with ViPT also leads to enhanced performance, with scores rising from 65.0, 61.6, and 52.4 to 66.4, 63.0, and 53.0.

Another competitor, $\rm{BD^2}$Track, as shown in Tab. \ref{tab:results-effectiveness}, employs a similar fusion strategy. 
However, its performance falls short, even lagging behind the simplest method Siamese+GMMT(U) by 1.7\% on SR.
When compared to ViPT*+GMMT or TBSI+GMMT, the performance gap is much larger.

Besides, on RGBT234, TBSI+GMMT continuously shows the best performance, reaching 64.7 and 87.9 on the SR and PR metrics, respectively.

On GTOT, we simultaneously display the overall performance together with the analysis on each attributes agianst CAT \cite{cat}, CMPP \cite{cmpp}, \cite{siamcda}, JMMAC \cite{JMMAC}, ADRNet \cite{adrnet}, MANet++ \cite{MANet}, HMFT \cite{hmft}, MacNet \cite{MaCNet}, APFNet \cite{apfnet}, and TBSI \cite{tbsi}, which are illustrated in Table. \ref{tab:gtotattribute}.
With the help of the proposed GMMT, TBSI+GMMT shows significantly improvement compared to the competitors.
Compared to the baseline method, TBSI, improvements of 2.6\% and 2.1\% are achieved on SR and PR, boosting the performance from 75.9 and 91.5 to 78.5 and 93.6, respectively, establishing a new state-of-the-art record on this benchmark.


To intuitively show the superiority of GMMT, the tracking results are displayed in Fig. \ref{fig:visall}, with additional visual comparisons available in the supplementary material.
In particular, for ViPT, the enhanced understanding provided by GMMT results in a noticeable improvement, as evident in the comparison between boxes colored in yellow and red.

\begin{table*}[h]
    \centering
        \caption{Results on LasHeR and RGBT234 benchmarks.}
    \resizebox{1\linewidth}{!}{ 
    \begin{tabular}{cccccccccccc}
    \toprule
     Benchmarks & Metrics & APFNet & DMCNet &HMFT & ProTrack & DFAT & HFRD & ViPT* & TBSI & ViPT*+GMMT & TBSI+GMMT \\
    \midrule
       \multirow{3}{*}{LasHeR} & PR $\uparrow$ &50.0&49.0&46.0&53.8&44.6&59.0 &65.0&69.2 &66.4 &70.7 \\
        & NPR $\uparrow$ &43.9&43.1&41.3&-& 40.0 &54.5&61.6&65.7&63.0& 67.0 \\
        & SR $\uparrow$ &36.2&35.5&32.6&42.0&33.6&46.4&52.4&55.6&53.0&56.6  \\
    \bottomrule
    \multirow{2}{*}{RGBT234} & PR $\uparrow$ &82.7&83.9&78.8&78.6&75.8&82.4 &83.5&87.1 &84.3&87.9 \\
        & SR $\uparrow$ &57.9&59.3&56.8&58.7&55.2&58.4&61.7&63.8&61.5&64.7  \\
        \bottomrule
    \end{tabular}
    }
    \label{tab:results-lasher}
\end{table*}

\begin{table*}[t]
\centering
\caption{Attribute-based Success/Precision score on GTOT dataset. The highest two scores are highlighted with \textbf{Bold} and \textit{Italics}. }
        \resizebox{1\linewidth}{!}{ 
\footnotesize
\label{tab:gtotattribute}
\begin{tabular}{c|cccccccccc|c}
\hline
    & CAT  & CMPP & SiamCDA & JMMAC  & ADRNet & MaNet++  & HMFT & MacNet  & APFNet  & TBSI&  TBSI+GMMT \\ \hline  
TC  & 71.0/90.0 & 72.9/\textbf{93.8} & 68.5/82.6  & 70.5/88.6  & 73.9/90.6   & 70.7/89.9    & 73.2/89.2 & 69.4/89.5    & 71.6/90.4  & 75.0/90.7 & \textbf{77.6}/\textit{92.6}    \\
OCC  & 69.2/89.9 & 71.6/\textbf{94.7}  & 69.4/82.2  & 68.7/84.0   & 69.9/87.9    & 70.1/89.0 & 72.2/88.1 & 68.5/88.2   &71.3/90.3 & 75.9/91.8 & \textbf{76.9}/\textit{92.7}  \\
LSV  & 68.0/85.0 & 70.0/91.2  & 74.8/91.5  & 74.6/90.3   & 70.8/85.5    & 69.3/86.6 & 75.4/89.1   & 67.1/84.9 & 71.2/87.8   & \textit{77.4}/\textit{93.8}  & \textbf{79.1}/\textbf{94.5}    \\
FM  & 65.4/83.9 & 68.6/91.7  & 74.8/91.5  & 74.6/90.3   & 70.8/85.5   & 69.3/86.6 & 75.4/89.1  & 67.1/84.9 & 71.2/87.8    & \textit{77.4}/\textit{93.8}    & \textbf{79.1}/\textbf{94.5}    \\
LI  & 72.3/89.2 & 74.3/92.4  & 76.4/92.4  &76.5/\textit{95.3}  & 76.2/91.9    & 73.1/91.7 & 76.9/94.3    & 72.9/90.0 & 74.8/91.4    & \textit{77.1}/94.1    & \textbf{80.1}/\textbf{96.5}    \\
SO  & 69.9/84.7 & 72.5/\textbf{98.1}  & 69.1/87.4  & \textit{73.8}/\textit{95.2} & 72.5/94.4 & 69.9/93.9 &71.6/92.5   &69.2/95.1 & 71.3/94.3    & 71.9/90.3   & \textbf{75.3}/92.8    \\
DEF  & 75.5/92.5 & \textbf{78.8}/94.6  & 72.7/87.9  & 76.2/\textbf{96.4} & 77.9/94.3 & 74.4/93.8 &74.8/94.0   &76.2/93.2 & 78.0/94.6    & 75.0/91.5   & \textit{78.4}/\textit{94.7}    \\
ALL  & 71.7/88.9 & 73.8/\textit{92.6}  & 73.2/87.7  & 73.2/90.2 & 73.9/90.5 & 72.3/90.1 &74.9/91.3   &71.2/88.6 & 73.7/90.5    & \textit{75.9}/91.5   & \textbf{78.5}/\textbf{93.6}    \\
\hline
\end{tabular}
}
\end{table*}

\subsection{RGB-D Extension}
To validate the generalisation of GMMT, we also implement it in the RGB-D tracking field, using ViPT-D as the baseline tracker.
ViPT-D is an extension of ViPT \cite{vipt} tailored for RGB-D data..
Initially, we run the official ViPT-D on the RGBD1K dataset, but we notice a performance gap compared to SPT \cite{rgbd1k}, which is currently the state-of-the-art method on this benchmark. 
As RGBD1K videos exhibit a higher diversity with more challenging factors compared to other RGB-D benchmarks, we retrain ViPT-D on the training split of RGBD1K. This retraining effort boosts the F-score from 46.2 to 50.6.
The further application of GMMT is based on this retrained variant, , which we denote as ViPT-D*.

The quantitative results are displayed in Table. \ref{tab:results-rgbd1k}, with the competitors being SPT, DDiMP \cite{dimp}, and DeT \cite{det}.
When the UNet is utilized as the embedding network of GMMT, a performance gain of 5.6\% is observed. 
This improvement becomes even more substantial when UNet is replaced by UViT, resulting in an F-score of 57.4.
Notably, in addition to the main metric, ViPT-D*+GMMT(V) surpasses ViPT-D* by 6.7\% and 7.0\% on PR and RE, demonstrating consistent enhancements across all metrics. These results highlight the superiority of GMMT.
Besides, although our baseline tracker ViPT-D* falls far behind SPT, which performs the best among all the competitors, a new state-of-the-art is built with the help of our GMMT. 
Furthermore, despite our baseline tracker, ViPT-D*, initially lagging behind SPT, which performs the best among all the competitors, we achieve a new state-of-the-art with the assistance of GMMT."

\begin{table}[t]
    \centering
    \caption{Results on RGBD1K benchmark.}
        \resizebox{1\linewidth}{!}{ 
    \begin{tabular}{ccccc}
    \toprule
      Method & PR $\uparrow$ & RE $\uparrow$ & F-score $\uparrow$ & $\Delta$\\
    \midrule
        DDiMP &55.7&53.4&54.5 &\\
        DeT&43.8&41.9&42.8 & \\
        SPT&54.5&57.8&56.1 &\\
    \toprule
        ViPT-D &45.3 & 47.2&46.2 &\\
        ViPT-D* &49.2&52.0&50.6 & \\
        ViPT-D*+GMMT(U)&54.7&57.9&56.2 &+5.6\%\\
        ViPT-D*+GMMT(V)&55.9&59.0&57.4&+6.8\%\\
    \bottomrule
    \end{tabular}
    }
    \label{tab:results-rgbd1k}
\end{table}

\subsection{Self-Analysis}
In this section, detailed discussions of our implementation are provided.
\begin{table}[t]
    \centering
    \caption{Effectiveness analysis on multiple baseline trackers and benchmarks. U and V are the abbreviation of UNet and UViT, respectively.}
    \resizebox{1\linewidth}{!}{ 
    \begin{tabular}{ccccccccccc}
    \toprule
       Benchmark & Method & PR$\uparrow$ & NPR$\uparrow$ & SR$\uparrow$ & $\Delta$ \\
    \midrule
        GTOT & Siamese &84.0&-&67.0&\\
        GTOT & Siamese+GMMT(U)&85.7&-&69.3& +2.3\% \\
    \midrule
        LasHeR & Siamese &50.9&47.4&39.8& \\
        LasHeR & Siamese+GMMT(U)&57.1&53.0&44.9& +5.1\%\\
    \midrule
        LasHeR & ViPT* &65.0&61.6&52.4&\\
        LasHeR & ViPT*+GMMT(U)&65.9&62.4&52.7&+0.3\%\\
        LasHeR & ViPT*+GMMT(V)&66.4&63.0&53.0&+0.6\%\\
    \midrule
        LasHeR & TBSI &69.2&65.7&55.6&\\
        LasHeR & TBSI+GMMT(U)&70.7&67.0&56.6&+1.0\%\\
        LasHeR & TBSI+GMMT(V)&70.5&66.6&56.3&+0.7\%\\
    \midrule
        LasHeR & $\rm{BD^2}$Track & 56.0 & - & 43.2 \\
    \bottomrule
    \end{tabular}
    }
    \label{tab:results-effectiveness}
\end{table}

\textbf{Implementation on Multiple Baseline Trackers:}
To prove our GMMT as a general fusion mechanism, various experiments are conducted on equipping multiple baseline trackers with our GMMT, including a self-designed Siamese tracker, the ViPT and TBSI.
On GTOT,  when using the UNet \cite{ddpm} as the embedding network $\mathcal{U}$, the SR of the Siamese baseline is enhanced from 67.0 to 69.3 through the combination of GMMT.
On LasHeR, the performance of our Siamese baseline can be significantly boosted from 39.8 to 44.9, with an increment of 5.1\%.
However, the improvements were relatively modest for ViPT* and TBSI, with gains of around 0.3\% and 1.0\%, respectively. 
We attribute this phenomenon to the fusion block in the baseline trackers.
In the self-designed Siamese baseline, the fusion block is lightweight.
It solely contains a convolutional block, leading to the multi-modal information being insufficiently fused.
Thus, prompted by GMMT, the performance climbs in a large step.
n contrast, ViPT* and TBSI already have well-established fusion processes for multi-modal information, leading to smaller performance increments. 
In ViPT*, the fusion process occurs in all the 12 self-attention blocks, and it only takes place in the 3$^{th}$, 6$^{th}$, and 9$^{th}$ blocks in TBSI.
Therefore, the improvement on ViPT* is slightly less than that on TBSI.
\textit{In conclusion, the worse the multi-modal information is fused in the baseline tracker, the more it can be boosted by our GMMT.}

The experiments on RGB-D tracking are also in line with this conclusion.
Based on the same ViPT baseline, a considerably larger improvement can be found in RGB-D benchmarks.
We owe this to the difference in the distinct characteristics of the input data.
Although RGB and T data have varying characteristics under various scenarios, they are both imaged based on electromagnetic waves.
However, the depth image reflects the distance signal of the surroundings, which has larger heterogeneity to the RGB data.
Consequently, the same fusion strategy employed in RGB-T data yields poorer results when applied to RGB-D data.
In other words, the multi-modal information in ViPT-D is more inadequately fused than ViPT, which gives reason for the larger enhancement observed in the RGBD1K benchmark.

\textbf{Learnable Network $\mathcal{U}$ in GMMT:}
To accomplish the GMMT, an learnable network $\mathcal{U}$ is necessarily introduced in the embedding GM.
In our implementation, two renowned networks, UNet \cite{ddpm} and UViT \cite{uvit}, are involved.
As their name suggests, both of them follow the U-shaped architecture introduced in supplementary material.
However, they differ in two significant aspects: the number of blocks and the detailed architecture of each block. UNet employs a convolution-based block, while UViT constructs its block using the transformer architecture.

The results are listed in Table. \ref{tab:results-effectiveness}.
With the UNet employed, on the SR metric, the performance of ViPT* reaches 52.7 and that of TBSI is improved from 55.6 to 56.6.
Replaced by the UViT, ViPT* performs better but the results of TBSI degrades slightly.
But in general, GMMT can boost the baseline methods on all three metrics consistently no matter which inner network is selected.

Details of the network architecture and the analysis of the number of blocks, $n$, are remained to the supplementary material.

\textbf{Analysis of the Generated Features:}

\begin{figure}[t]
\centering
	\setcounter {subfigure} {0} a){
		\includegraphics[scale=0.22]{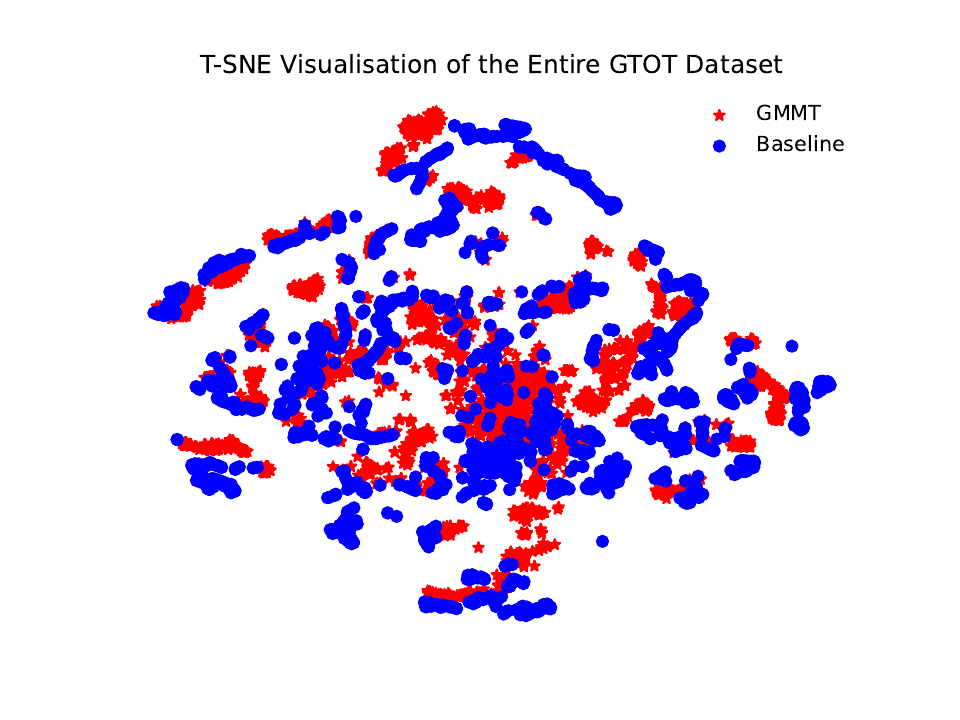}}
	\setcounter {subfigure} {0} b){
		\includegraphics[scale=0.22]{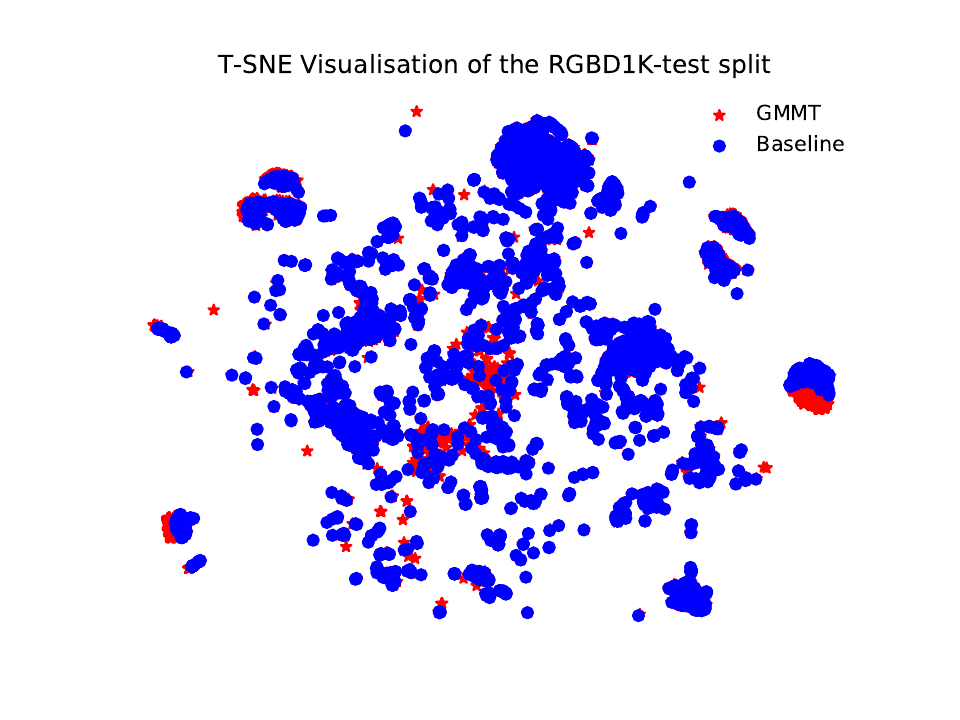}}
	\\
	\setcounter {subfigure} {0} c){
		\includegraphics[scale=0.31]{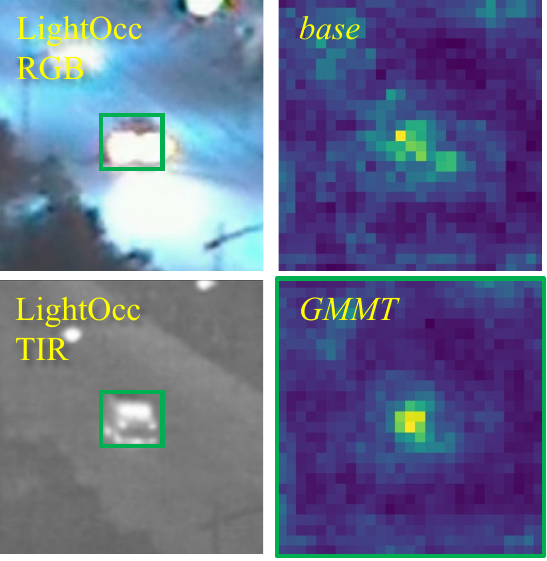}}
  	\setcounter {subfigure} {0} d){
		\includegraphics[scale=0.31]{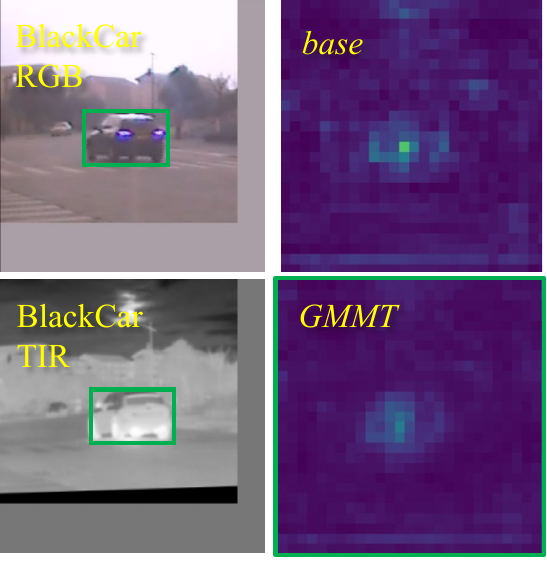}}
\caption{Visualisation of the feature embeddings.}
\label{fig:feature-difference}
\end{figure}

\begin{figure}[t]
\centering
\includegraphics[width=1.0\linewidth]{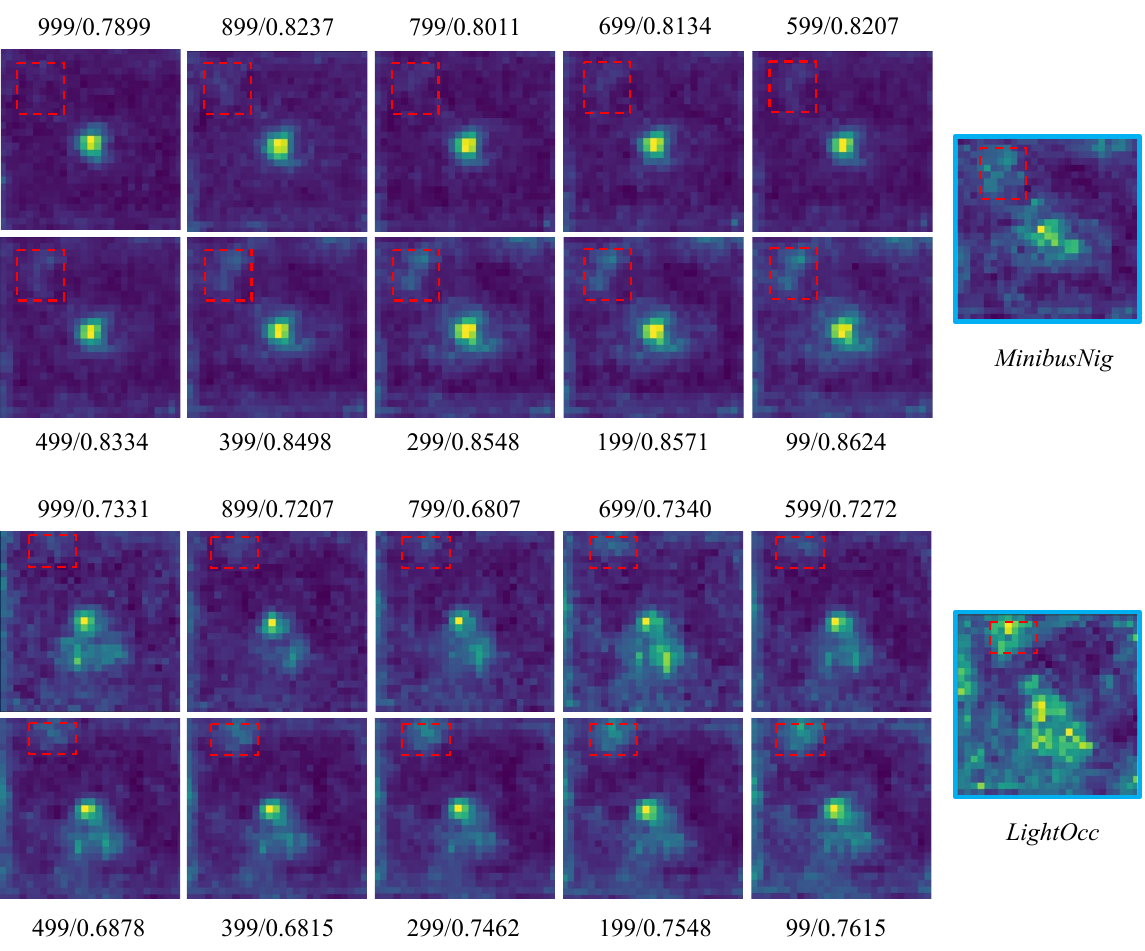}
\caption{Analysis of diffusion Steps. The content below each feature map is the time step and its structural similarity between the original fused features on the right side.}
\label{fig:steps}
\end{figure}

To verify the superiority of GMMT intuitively, the generated features are visualised in Fig. \ref{fig:feature-difference}.
Specifically, it is demonstrated globally and locally.

Through the global description, the effectiveness of the generated features is proved.
The t-sne tool is used to exhibit thousands of the original and generated features.
Fig. \ref{fig:feature-difference}(a) and Fig. \ref{fig:feature-difference}(b) are the statistical analysis on GTOT and RGBD1K datasets.
In these two graphs, the original features are marked with blue circles and the generated ones are highlighted by the red pentagram.
Apparently, the clusters of the generated and the original fused features are highly overlapped.
This overlap indicates that they occupy the same semantic space and share similar properties. 
Consequently, we believe the generated features are capable of supporting the tracking task, just as the original ones.

After the effectiveness is proved, the superiority is demonstrated by the local description.
We provide the visualisation of the feature maps before and after combining GMMT in Fig. \ref{fig:feature-difference}(c) and Fig. \ref{fig:feature-difference}(d).
Two samples from GTOT are displayed.
The left one is from the video $LightOCC$ and another is from $BlackCar$.
In these two instances, the targets locate in the centre because they are centre-cropped before sending into the network.
Therefore, the ideal feature maps should exhibit a strong response in the central target area while suppressing background regions.
In $LightOCC$, the visualisation from GMMT has a higher response in the target area, and the background is clearer because the region with extreme illumination in the RGB modality better discarded.
In $BlackCar$, both feature maps are focused on the key position, but the background noise in GMMT is better suppressed.
Based on the analysis of these two video samples, we attribute the superiority of GMMT to its ability to generate more discriminative features. Further visualizations and analyses are available in the supplementary material, where consistent conclusions are drawn.

The reason for producing better features is attributed to the generative training paradigm.
During training, a random noise together with the RGB and TIR features form the input of the network $\mathcal{U}$.
In this setup, $\mathcal{U}$ needs to effectively understand and extract crucial cues from both $f_{rgb}$ and $f_{tir}$ to successfully perform the information fusion task.
As a result, compared to a network trained with a purely discriminative paradigm, our approach encourages $\mathcal{U}$ to better extract important information from each modality. This, in turn, leads to fused features with enhanced discrimination, making them more suitable for challenging tracking tasks that involve diverse and complex environmental conditions.


\textbf{Analysis of Diffusion Steps:}
Different from the typical fusion blocks, our GMMT(DM) can be recursively executed.
Fig. \ref{fig:steps} gives the visualisation of $s=10$ steps (from 999 to 99, in a reverse manner) when T is set to 1000.
It can be seen that with $s$ becomes larger, the generated features are more similar to the original fused features $fused$.
Unexpectedly, the noise in $fused$ is also better recovered as shown in Fig. \ref{fig:steps}(a) and (b).
The similarity is quantified by the structural similarity between the generated feature map $fused^*$ and $fused$.
In general, the similarity becomes larger with more reverse diffusion steps, and the last steps in both instances have the largest scores. 
This means the superiority of GMMT is disappearing gradually when $s$ goes larger.
The quantitative results and the corresponding analysis are displayed in the supplementary material, and no positive relationship is observed between the performance and $s$.
Besides, more steps also cost more computational resources and time.
Therefore, $s$ equals 1 in our method for efficiency, reaching 16 frames per second.

\textbf{Analysis of $\lambda$:}
$\lambda$ is a crucial factor banding the tracking and generation tasks.
Thus, the analysis on it is conducted and exhibited in the supplementary material, with $\lambda$ valued from (1,2,3,5,10,100).
The conclusion is that all the variants perform better than the baseline method, which demonstrate the superiority of GMMT.
Additionally, when $\lambda$=100, the performance is lightly better than the baseline.
This indicates that $\lambda$ should not be a large value, leading to a small influence of the tracking loss, and, furthermore, the strong supervision of the tracking task is crucial and should not be ignored.

\section{Conclusion}
This paper proposes a novel generative-based fusion mechanism for multi-modal tracking, known as GMMT.
Its effectiveness has been demonstrated through the implementation on multiple tracking baselines, the evaluation on multiple challenging benchmarks, as well as two multi-modal tracking tasks.
Enhanced by our GMMT, a new state-of-the-art is built on the challenging LasHeR and RGBD1K benchmarks.
Furthermore, through the intuitive visualisation, we attribute its superiority to the noisy training paradigm, which understands and preserves the discriminative clues from each modality to the fused features.
Additionally, GMMT tends to yield larger improvements when applied to baseline methods with rough information fusion processes.

\section*{Acknowledgement}

This work was supported in part by the National Natural Science Foundation of China (62020106012, U1836218, 61672265, 62106089), the 111 Project of Ministry of Education of China (B12018), and the Engineering and Physical Sciences Research Council (EPSRC) (EP/N007743/1, MURI/EPSRC/DSTL, EP/R018456/1).

\bibliographystyle{IEEEtran} 
\bibliography{ref} 

\clearpage
\section{Appendix}
To deliver a precise introduction of the proposed generative-based fusion mechanism for multi-modal tracking (GMMT), some of the details are exhibited in this supplementary material.
\begin{itemize}
 	\item \textbf{A.} The introduction of the discriminator, $D$, in the conditional generative adversarial network (CGAN).
        \item \textbf{B.} Benchmarks and metrics.
        \item \textbf{C.} More details when implementing GMMT.
        \item \textbf{D.} Intuitive comparison between the original fusion block and the proposed GMMT.
	\item \textbf{E.} More visualisation of the tracking results.
        \item \textbf{F.} The quantitative results produced during the analysis of $n$, the number blocks in the U-shape network.
        \item \textbf{G.} The quantitative results produced during the analysis of $s$, the number of steps the reverse diffusion process takes.
        \item \textbf{H.} Analysis of $\lambda$.
\end{itemize}

\subsection{A. Discriminator in CGAN}

\begin{figure}[t]
\centering
\includegraphics[width=0.5\textwidth]{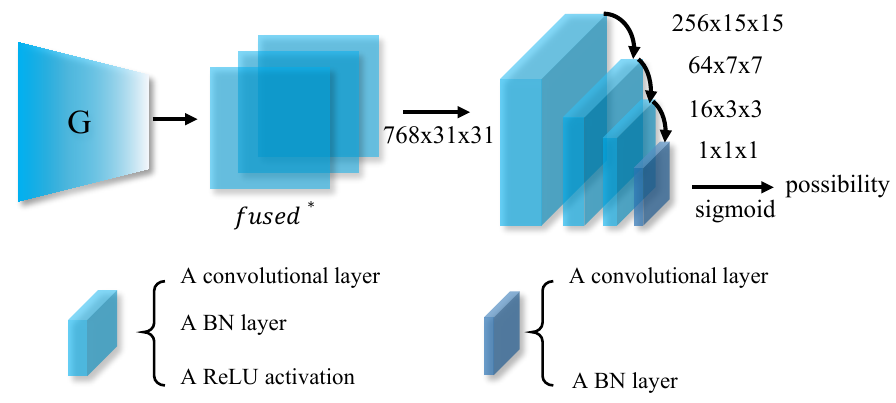}
\caption{Architecture of the discriminator embedded in our CGAN-based GMMT.}
\label{fig:cgan-D}
\end{figure}

Since the generator $G$ kept the same for all the variants, it remains in the next section and only the discriminator $D$ is introduced in this part, which is shown in Fig. \ref{fig:cgan-D}.
$D$ receives the output of $G$ and four convolutional blocks are embedded for dimension reduction.
Later, a sigmoid activation is employed to transfer the output to the interval [0,1].
The output serves the possibility to be a real sample.
Each convolutional block contains a convolutional layer, a batch normalisation layer, and a ReLu activation.
Notably, there is no activation in the last convolution block. 

\subsection{B. Benchmark and Metrics}
The effectiveness of GMMT is mainly verified on GTOT \cite{gtot} and  LasHeR \cite{lasher}

GTOT is an early published RGB-T dataset, including 7.8K image pairs.
The evaluation metrics are precision rate (PR) and success rate (SR).
PR measures the percentage of frames with the distance between centres of the predicted and ground truth bounding box below a threshold, 5 in this benchmark.
SR represents the ratio of frames being tracked with the overlap between the predicted and ground truth bounding box above zero.

LasHeR is a large and widely-used benchmark in the RGB-T field, and its testing split consists of 245 video pairs.
PR, SR and the normalised precision rate (NPR) are used for benchmarking.
NPR \cite{trackingnet} is a modified version of PR since PR can be easily affected the image resolution and the size of the ground truth bounding box.
It should be noted that the threshold of PR in LasHeR is 20.

To demonstrate that our GMMT is a general fusion mechanism, experiments on RGBD1K are also conducted.
Its testing split involves 50 video pairs.
In these videos, the object can be absent, and therefore, the recall (RE) is employed to calculate the rate of the object being successfully tracked.
The definition of being successfully tracked means the overlap between the predicted and ground truth is above zero.
Later, F-score, a comprehensive metric, is further introduced by taking both PR and RE into account.

\subsection{C. Implementing Details}
\begin{figure*}[t]
\centering
\includegraphics[width=1.0\textwidth]{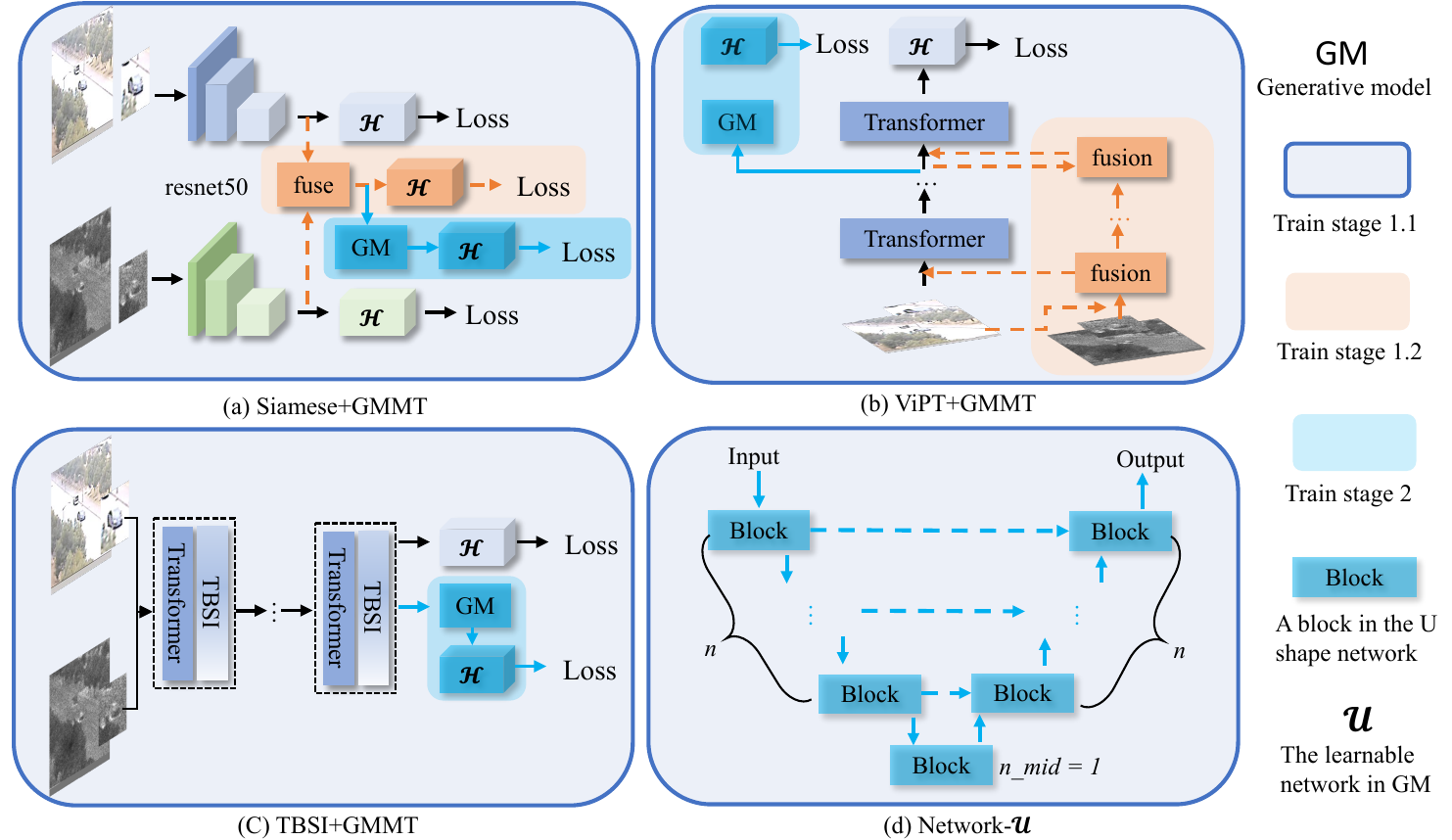}
\caption{The implementation of GMMT on multiple baseline trackers (a)Siamese, (b)ViPT, (c)TBSI, and (d) the architecture of the learnable network $\mathcal{U}$ in GMMT. We follow the official implementation of ViPT and TBSI while the Siamese baseline is a self-designed one.}
\label{fig:methods}
\end{figure*}

The experimental details are introduced in this part.
Basically, the effectiveness of GMMT is demonstrated on three baseline methods, \ie a self-designed Siamese tracker, the ViPT \cite{vipt} and TBSI \cite{tbsi}, which are sequentially introduced in the following paragraphs.

The first selected baseline is a Siamese tracker constructed based on the advanced RGB tracker SiamBAN \cite{siamban}.
SiamBAN consists of a ResNet50 \cite{resnet50} backbone, a Neck block, and three tracking heads.
In our design, only the output of the third residual layer of ResNet50 is maintained for efficiency, and the number of tracking heads is also reduced to 1.
As a multi-modal tracker, the above baseline is duplicated once, which is further applied to the TIR data.
The overall architecture is shown in Fig. \ref{fig:methods}(a).
As Fig. \ref{fig:methods}(a) shows, the SiamBAN baseline is trained at first.
Later, the fusion block (a convolutional layer) as well as its corresponding tracking head are optimised with other parts frozen.
Our generative model (GM) is embedded in our GMMT and a tracking head is trained at last.

The second baseline tracker is ViPT, as shown in Fig. \ref{fig:methods}{b}. 
Basically, an offline-trained model is provided by the authors.
However, it performs 51.9 on the SR while 52.5 in the published manuscript, which leads to the retraining of the fusion blocks (also termed prompt layers).
After that, our GMMT and the extra tracking head are appended and optimised.

With the least modification, the application on the third baseline, TBSI, is illustrated in Fig. \ref{fig:methods}(c).
The training procedure of GMMT is activated after the parameters of the official-provided model are frozen.

Notably, our GMMT is applied to the features from search image patches, and almost all the crucial configurations during the training procedure are displayed in Table. \ref{tab:configurations}.

After the training of GMMT, it is then evaluated on the multi-modal benchmarks.
To present the difference between the typical fusion method and GMMT, the pseudo-code is programmed in Table. \ref{tab:pesudo-code}.
Effortlessly, the GMMT can be distinguished from the typical fusion methods.
Firstly, the original discriminative-based fusion block is abandoned and, instead,  a generative-based fusion block is activated.
Secondly, our GMMT can be executed iteratively, which property is unseen in the existing fusion methods.

\begin{table}[t]
    \setlength{\tabcolsep}{1.0mm}
{
    \centering
    \caption{Pseudo code of GMMT.}
    \begin{tabular}{cl}
    \toprule
       & \textbf{Pseudo code when testing}\\
    \midrule
       \textbf{Input:}&  Current features from each modality $f_{rgb,t}$, $f_{tir,t}$, \\
                    &  Noise $z$,Tracking Head $\mathcal{H}$, Generator $G$ \\
                    &  Feature extractor and fusion block $\mathcal{F}$\\
    \midrule
       \textbf{Track:}& IF Type == 'Typical': \\
                    &  \qquad $fused$ = $\mathcal{F}$($f_{rgb}$, $f_{tir}$) \\
                    &  ELIF Type == 'GMMT': \\
                    &  \qquad while iteration \\
                    &  \qquad \qquad $fused^*$ = $\mathcal{G}$($z$, $f_{rgb}$, $f_{tir}$) \\
                    &  \qquad \qquad $z$ = $fused^*$\\
                    &  \qquad $fused$ = $fused^*$ \\
    \midrule
       \textbf{Output:}&  Prediction of current frame $\mathcal{P}$=$\mathcal{H}$($fused$)\\
    \bottomrule
    \end{tabular}
    }
    \label{tab:pesudo-code}
\end{table}

\begin{table*}[t]
\centering
\caption{Experimental configurations on three baseline methods during the multi-stage training scheme. For each hyper-parameter, there are three rows, which are held for the Siamese baseline, ViPT, and TBSI, sequentially.}
\begin{tabular}{|c|ccc|}
\hline
Configurations   & Stage 1.1                                                                                                     & Stage 1.2                                                                                                     & Stage 2                                                                                                          \\ \hline
Batchsize        & \multicolumn{3}{c|}{8}                                                                                                                                                                                                                                                                                                                           \\ \hline
                 & \multicolumn{1}{c|}{-}                                                                                        & \multicolumn{2}{c|}{32}                                                                                                                                                                                                          \\ \hline
                 & \multicolumn{1}{c|}{-}                                                                                        & \multicolumn{1}{c|}{-}                                                                                        & \multicolumn{1}{c|}{32}                                                                                          \\ \hline
Epoch            & \multicolumn{2}{c|}{20}                                                                                                                                                                                                       & \multicolumn{1}{c|}{100}                                                                                         \\ \hline
                 & \multicolumn{1}{c|}{-}                                                                                        & \multicolumn{2}{c|}{100}                                                                                                                                                                                                         \\ \hline
                 & \multicolumn{1}{c|}{-}                                                                                        & \multicolumn{1}{c|}{-}                                                                                        & \multicolumn{1}{c|}{100}                                                                                         \\ \hline
Learnable blocks & \multicolumn{1}{c|}{Backbone,Neck, RPN}                                                        & \multicolumn{1}{c|}{Fusion block}                                                                             & \multicolumn{1}{c|}{GMMT}                                                                                        \\ \hline
                 & \multicolumn{1}{c|}{-}                                                                                        & \multicolumn{1}{c|}{Promt layers}                                                                             & \multicolumn{1}{c|}{GMMT}                                                                                        \\ \hline
                 & \multicolumn{1}{c|}{-}                                                                                        & \multicolumn{1}{c|}{-}                                                                                        & \multicolumn{1}{c|}{GMMT}                                                                                        \\ \hline
Optimiser        & \multicolumn{1}{c|}{SGD}                                                                                      & \multicolumn{1}{c|}{SGD}                                                                                      & \multicolumn{1}{c|}{SGD}                                                                                         \\ \hline
                 & \multicolumn{1}{c|}{-}                                                                                        & \multicolumn{1}{c|}{ADAMW}                                                                                      & \multicolumn{1}{c|}{SGD}                                                                                         \\ \hline
                 & \multicolumn{1}{c|}{-}                                                                                        & \multicolumn{1}{c|}{-}                                                                                        & \multicolumn{1}{c|}{SGD}                                                                                         \\ \hline
Base LR          & \multicolumn{3}{c|}{0.005}                                                                                                                                                                                                                                                                                                                       \\ \hline
                 & \multicolumn{1}{c|}{-}                                                                                        & \multicolumn{1}{c|}{0.0004}                                                                                   & \multicolumn{1}{c|}{0.005}                                                                                       \\ \hline
                 & \multicolumn{1}{c|}{-}                                                                                        & \multicolumn{1}{c|}{-}                                                                                        & \multicolumn{1}{c|}{0.005}                                                                                       \\ \hline
Learning rate    & \multicolumn{1}{c|}{\begin{tabular}[c]{@{}c@{}}1$\sim$5:0.001-0.005, \\ 6$\sim$20:0.005-0.00005\end{tabular}} & \multicolumn{1}{c|}{\begin{tabular}[c]{@{}c@{}}1$\sim$5:0.001-0.005, \\ 6$\sim$20:0.005-0.00005\end{tabular}} & \multicolumn{1}{c|}{\begin{tabular}[c]{@{}c@{}}1$\sim$20:0.001-0.005, \\ 21$\sim$100:0.005-0.00005\end{tabular}} \\ \hline
                 & \multicolumn{1}{c|}{-}                                                                                        & \multicolumn{1}{c|}{-}                                                                                    & \multicolumn{1}{c|}{\begin{tabular}[c]{@{}c@{}}1$\sim$20:0.001-0.005, \\ 21$\sim$100:0.005-0.00005\end{tabular}} \\ \hline
                 & \multicolumn{1}{c|}{-}                                                                                        & \multicolumn{1}{c|}{-}                                                                                        & \multicolumn{1}{c|}{\begin{tabular}[c]{@{}c@{}}1$\sim$20:0.001-0.005, \\ 21$\sim$100:0.005-0.00005\end{tabular}} \\ \hline
Weight decay     & \multicolumn{3}{c|}{0.0001}                                                                                                                                                                                                                                                                                                                      \\ \hline
                 & \multicolumn{1}{c|}{-}                                                                                        & \multicolumn{2}{c|}{0.0001}                                                                                                                                                                                                      \\ \hline
                 & \multicolumn{1}{c|}{-}                                                                                        & \multicolumn{1}{c|}{-}                                                                                        & \multicolumn{1}{c|}{0.0001}                                                                                      \\ \hline
Momentum         & \multicolumn{3}{c|}{0.9}                                                                                                                                                                                                                                                                                                                         \\ \hline
                 & \multicolumn{1}{c|}{-}                                                                                        & \multicolumn{1}{c|}{No}                                                                                       & \multicolumn{1}{c|}{0.9}                                                                                         \\ \hline
                 & \multicolumn{1}{c|}{-}                                                                                        & \multicolumn{1}{c|}{-}                                                                                        & \multicolumn{1}{c|}{0.9}                                                                                         \\ \hline
T                & \multicolumn{1}{c|}{-}                                                                                        & \multicolumn{1}{c|}{-}                                                                                        & \multicolumn{1}{c|}{1000}                                                                                        \\ \hline
Samper           & \multicolumn{1}{c|}{-}                                                                                        & \multicolumn{1}{c|}{-}                                                                                        & \multicolumn{1}{c|}{DDIM}                                                                                        \\ \hline
\end{tabular}
\label{tab:configurations}
\end{table*}

\subsection{D. Intuitive Comparison for GMMT}

\begin{figure*}[h]
\centering
\includegraphics[width=1.0\textwidth]{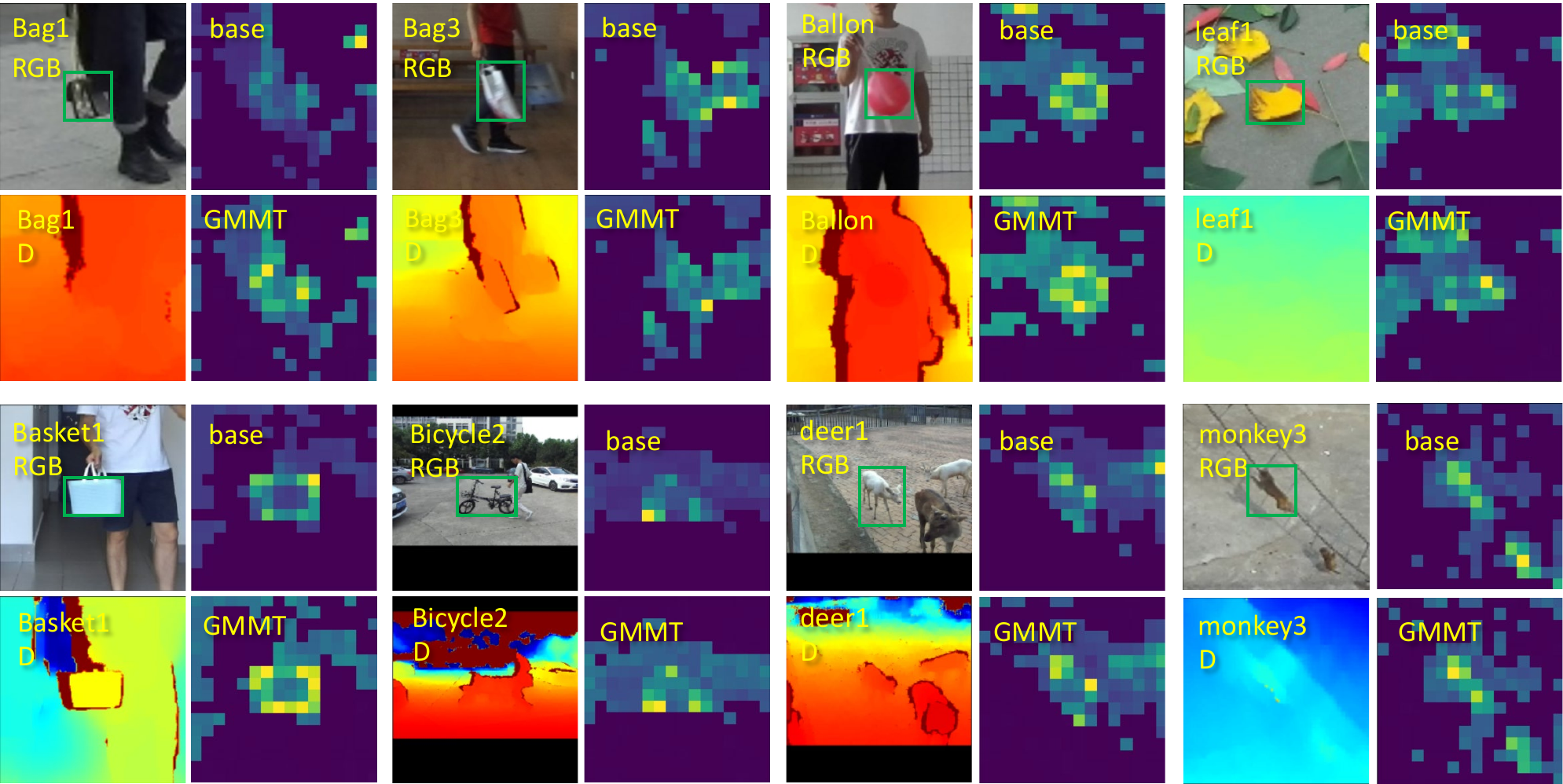}
\caption{Visualisation of the feature embeddings on the challenging RGBD1K benchmark.}
\label{fig:features}
\end{figure*}

To exhibit the superiority of the proposed GMMT, the comparison between the features with and without GMMT is intuitively provided in Fig. \ref{fig:features}, with all examples sampled from the RGBD1K benchmark.
In the instances sampled from $Basket1, Bicycle2$, the response in the target regions are significantly enhanced.
In other samples, the highest responses appear in the incorrect places, while they are corrected with the equipment of our GMMT.
This indicates that the features generated by GMMT have better discrimination, as well as a higher possibility to predict compact bounding boxes.

\begin{table*}[h]
    \centering
    \caption{Quantitative analysis of diffusion steps.}
    \setlength{\tabcolsep}{2.5mm}
{
    \begin{tabular}{ccccccccccccccc}
    \toprule
       s & 1 & 2 & 3 & 4 & 5 & 6 &7 & 8 & 9 & 15&20&30&40 \\
    \midrule
        PR$\uparrow$ & 85.7 & 84.5 &85.0&85.2&84.9&85.1&83.9&84.9&83.8&84.4&85.3&84.7&85.6\\
        SR$\uparrow$ & 69.3 & 68.4 &68.5&68.5&68.5&68.4&67.9&68.4&68.0&68.1&68.7&68.4&68.9\\
    \bottomrule
    \end{tabular}
    \label{tab:results-steps}
    }
\end{table*}



\subsection{E. Tracking Results}

\begin{figure*}[t]
\centering
\includegraphics[width=1.0\textwidth]{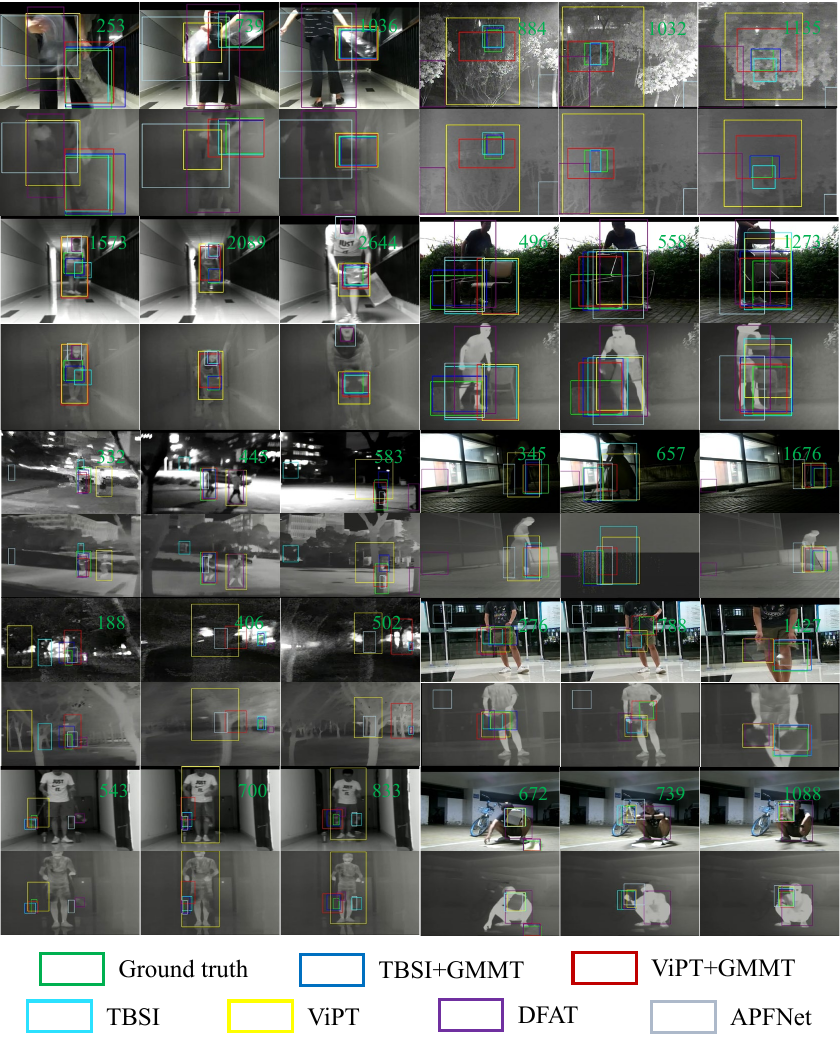}
\caption{Qualitative results on LasHeR. The exhibited image pairs are sampled from video $lefthyalinepaperfrontpants$, $hyalinepaperfrontface$, $boytakingbasketballfollowing$, $darktreesboy$, $leftbottle2hang$, $boyride2path$, $leftchair$, $broom$, $large$, $leftexcersicebookyellow$, which are introduced in a top-down and left-right way. }
\label{fig:vis-all}
\end{figure*}

More intuitive tracking results are provided in Fig. \ref{fig:vis-all}.
Under more challenging scenarios, the TBSI+GMMT and ViPT+GMMT still perform significantly better than the baseline methods, TBSI and ViPT, respectively, which further demonstrates the superiority of the proposed GMMT.

\begin{figure}[h]
\centering
\includegraphics[width=0.5\textwidth]{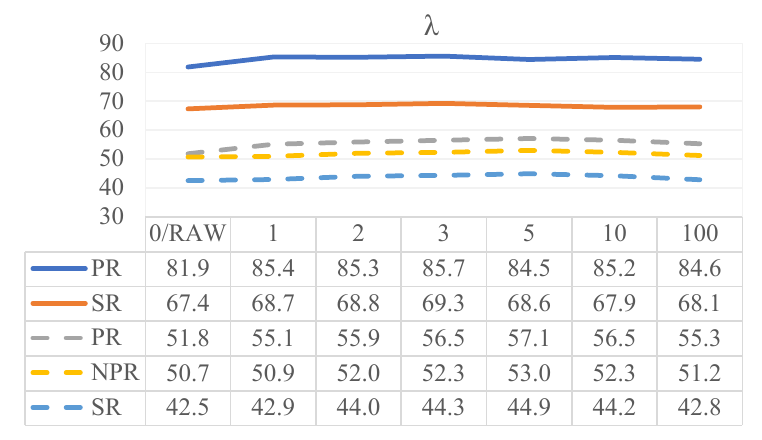}
\caption{Analysis of $\lambda$.}
\label{fig:lambda}
\end{figure}

\subsection{F. Number of Blocks}

\begin{table}[h]
    \centering
    \caption{Analysis of $n$ in the U-shaped network. OoM is the abbreviation of out of memory.}
    \setlength{\tabcolsep}{0.8mm}
{
    \begin{tabular}{cccccccccc}
    \toprule
    UViT \\
       n & 1 & 2 & 3 & 4 & 5 & 6 &7 & 8 & 9 \\
    \midrule
        PR$\uparrow$ & 51.8 & 52.5 &52.8&53.8&53.3&53.5&53.2&55.9&OoM\\
        RE$\uparrow$ & 54.6 & 55.4 &55.7&57.0&56.3&56.5&56.4&59.0&OoM\\
        F-score$\uparrow$ & 53.2 & 53.9 &54.2&55.4&54.8&54.9&54.7&57.4&OoM\\
    \bottomrule
    UNet \\
       n & 3& & 6& & 9& & 12 && 15 \\
    \midrule
        PR$\uparrow$ & 53.5 &  &54.0&&54.7&&53.5&&OoM\\
        RE$\uparrow$ & 56.4 &  &57.1&&57.9&&56.6&&OoM\\
        F-score$\uparrow$ & 55.0 &  &55.5&&56.2&&55.0&&OoM\\
    \bottomrule
    \end{tabular}
    \label{tab:results-blocks}
    }
\end{table}

As shown in Fig. \ref{fig:methods}(d), the depth of the embedding U-shape network is determined by a hyper-parameter $n$, which represents the number of blocks.
The exploration of network depth is a long-lasting topic in the deep learning era.
Therefore, the value of $n$ is investigated in this part, and Table. \ref{tab:results-blocks} gives the quantitative results on RGBD1K, with both the UNet \cite{ddim} and UViT \cite{uvit} are involved.
When using UViT, the best performance is achieved when $n$ equals 8, achieving 57.4 on F-score.
When UNet is selected, the best performance is 1.2\% worse than UViT, reaching 56.2 on F-score.
Compared to the state-of-the-art tracker SPT, the performance of these two variants is better, especially the UViT-based GMMT, owning an increment of 1.3\%.

Besides, the performance of our baseline method ViPT* is 49.2, 52.0, and 50.6 on PR, RE and F-score, respectively.
That is to say, no matter which inner network is chosen and how many blocks are stacked, our GMMT can consistently boost the tracking performance.

As to the architecture of each block in UNet and UViT, it is beyond the scope of this manuscript, and please refer to \cite{ddim} and \cite{uvit}, respectively.

\subsection{G. Analysis of $\lambda$}
$\lambda$ is a factor banding the tracking and generation tasks.
Fig. \ref{fig:lambda} shows the analysis on GTOT (solid) and LasHeR (dashed) benchmarks, with $\lambda$ chosen from (0,1,2,3,5,10,100).
0 represents the generative loss is inactivated and the network $\mathcal{U}$ is optimized by the $Loss_{track}$, which is inherited from the baseline methods.
When $\lambda$ is a nonzero value, stable improvements can be found on all the metrics.
However, the performance suffers a heavy degradation when $\lambda$=100.
At this time, the $Loss_{track}$ is too small and its influence on $\mathcal{U}$ is negligible.
Therefore, the above phenomenon indicates that the strong supervision of the tracking task is crucial and should not be ignored.

\subsection{H. Number of Steps}

Table. \ref{tab:results-steps} displays the quantitative results during the analysis of reverse diffusion steps $s$.
From this table, no positive correlation between the performance and $s$ is observed.
Meanwhile, a larger $s$ will cost more time and computational resources, which harms the efficiency significantly.
Therefore, $s$ is set to 1.

\end{document}